\documentclass[11pt,a4paper]{article}
\usepackage[hyperref]{emnlp2018}
\usepackage{times}
\usepackage{latexsym}

\usepackage{natbib}
\usepackage{url}
\usepackage{graphicx, wrapfig, graphics}
\usepackage{amsfonts, amssymb, amsmath}
\usepackage{subfig}
\usepackage{multirow}
\usepackage{soul}
\usepackage{enumitem}
\usepackage{color}
\usepackage{booktabs}
\usepackage{upgreek}
\usepackage[hang]{footmisc}
\usepackage{tikz}
\usepackage[linesnumbered,ruled]{algorithm2e}
\usepackage{float}
\usepackage{tabulary}

\aclfinalcopy % Uncomment this line for the final submission
\def\aclpaperid{413} %  Enter the acl Paper ID here

\title{Paraphrase Generation with Deep Reinforcement Learning}

\author{\sf Zichao Li$^1$, Xin Jiang$^1$, Lifeng Shang$^1$, Hang Li$^2$\\
$^1$Noah's Ark Lab, Huawei Technologies\\
{\tt \{li.zichao, jiang.xin, shang.lifeng\}@huawei.com}\\
$^2$Toutiao AI Lab\\
{\tt lihang.lh@bytedance.com}}

\date{}

\newcommand{\MP}{M_{\phi}}
\newcommand{\GT}{G_{\theta}}

\def\endthebibliography{%
  \def\@noitemerr{\@latex@warning{Empty `thebibliography' environment}}%
  \endlist
}

\makeatletter

\newcommand{\Rmnum}[1]{\expandafter\@slowromancap\romannumeral #1@}

\makeatother
\begin{document}

\maketitle

\begin{abstract}
Automatic generation of paraphrases from a given sentence is an important yet challenging task in natural language processing (NLP). In this paper, we present a deep reinforcement learning approach to paraphrase generation. Specifically, we propose a new framework for the task, which consists of a \textit{generator} and an \textit{evaluator}, both of which are learned from data. The generator, built as a sequence-to-sequence learning model, can produce paraphrases given a sentence. The evaluator, constructed as a deep matching model, can judge whether two sentences are paraphrases of each other. The generator is first trained by deep learning and then further fine-tuned by reinforcement learning in which the reward is given by the evaluator. For the learning of the evaluator, we propose two methods based on \textit{supervised learning} and \textit{inverse reinforcement learning} respectively, depending on the type of available training data. Experimental results on two datasets demonstrate the proposed models (the generators) can produce more accurate paraphrases and outperform the state-of-the-art methods in paraphrase generation in both automatic evaluation and human evaluation.
\end{abstract}

\section{Introduction}
Paraphrases refer to texts that convey the same meaning but with different expressions. For example, ``\textit{how far is Earth from Sun}'', ``\textit{what is the distance between Sun and Earth}'' are paraphrases. Paraphrase generation refers to a task in which given a sentence the system creates paraphrases of it. Paraphrase generation is an important task in NLP, which can be a key technology in many applications such as retrieval based question answering, semantic parsing, query reformulation in web search, data augmentation for dialogue system. However, due to the complexity of natural language, automatically generating accurate and diverse paraphrases is still very challenging.
Traditional symbolic approaches to paraphrase generation include rule-based methods~\citep{mckeown1983paraphrasing}, thesaurus-based methods~\citep{bolshakov2004synonymous,kauchak2006paraphrasing}, grammar-based methods~\citep{narayan2016paraphrase}, and statistical machine translation (SMT) based methods~\citep{quirk2004monolingual,zhao2008combining,zhao2009application}.

Recently, neural network based sequence-to-sequence (Seq2Seq) learning has made remarkable success in various NLP tasks, including machine translation, short-text conversation, text summarization, and question answering (e.g., ~\citet{cho2014learning,wu2016google,shang15neural,vinyals2015neural,rush2015neural,yin2015neural}). Paraphrase generation can naturally be formulated as a Seq2Seq problem~\citep{cao2017joint,prakash2016neural,gupta2017deep,su2017cross}.
The main challenge in paraphrase generation lies in the definition of the evaluation measure. Ideally the measure is able to calculate the semantic similarity between a generated paraphrase and the given sentence. In a straightforward application of Seq2Seq to paraphrase generation one would make use of cross entropy as evaluation measure, which can only be a loose approximation of semantic similarity. To tackle this problem, ~\citet{ranzato2015sequence} propose employing reinforcement learning (RL) to guide the training of Seq2Seq and using lexical-based measures such as BLEU~\citep{papineni2002bleu} and ROUGE~\citep{lin2004rouge} as a reward function. However, these lexical measures may not perfectly represent semantic similarity. It is likely that a correctly generated sequence gets a low ROUGE score due to lexical mismatch. For instance, an input sentence ``\textit{how far is Earth from Sun}'' can be paraphrased as ``\textit{what is the distance between Sun and Earth}'', but it will get a very low ROUGE score given ``\textit{how many miles is it from Earth to Sun}'' as a reference.

In this work, we propose taking a data-driven approach to train a model that can conduct evaluation in learning for paraphrasing generation. The framework contains two modules, a generator (for paraphrase generation) and an evaluator (for paraphrase evaluation). The generator is a Seq2Seq learning model with attention and copy mechanism~\citep{bahdanau2014neural,see2017get}, which is first trained with cross entropy loss and then fine-tuned by using policy gradient with supervisions from the evaluator as rewards. The evaluator is a deep matching model, specifically a decomposable attention model~\citep{parikh2016decomposable}, which can be trained by supervised learning (SL) when both positive and negative examples are available as training data, or by inverse reinforcement learning (IRL) with outputs from the generator as supervisions when only positive examples are available. In the latter setting, for the training of evaluator using IRL, we develop a novel algorithm based on max-margin IRL principle~\citep{ratliff2006maximum}. Moreover, the generator can be further trained with non-parallel data, which is particularly effective when the amount of parallel data is small.

We evaluate the effectiveness of our approach using two real-world datasets (Quora question pairs and Twitter URL
paraphrase corpus) and we conduct both automatic and human assessments. We find that the evaluator trained by our methods can provide accurate supervisions to the generator, and thus further improve the accuracies of the generator. The experimental results indicate that our models can achieve significantly better performances than the existing neural network based methods.

It should be noted that the proposed approach is not limited to paraphrase generation and can be readily applied into other sequence-to-sequence tasks such as machine translation and generation based single turn dialogue. Our technical contribution in this work is of three-fold:

\begin{enumerate}[leftmargin=*, topsep=-0.1pt]
\setlength{\itemsep}{0pt}
\setlength{\parskip}{0pt}
\item We introduce the generator-evaluator framework for paraphrase generation, or in general, sequence-to-sequence learning.
\item We propose two approaches to train the evaluator, i.e., supervised learning and inverse reinforcement learning.
\item In the above framework, we develop several techniques for learning of the generator and evaluator.
\end{enumerate}

Section \ref{sec:bg} defines the models of generator and evaluator. In section \ref{sec:models}, we formalize the problem of learning the models of generator and evaluator. In section \ref{sec:exp}, we report our experimental results. In section \ref{seq:relate}, we introduce related work.

\section{Models}\label{sec:bg}

This section explains our framework for paraphrase generation, containing two models, the generator and evaluator.

\subsection{Problem and Framework}
Given an input sequence of words $X=[x_1,\ldots,x_S]$ with length $S$, we aim to generate an output sequence of words $Y=[y_1,\ldots,y_T]$ with length $T$ that has the same meaning as $X$. We denote the pair of sentences in paraphrasing as $(X, Y)$. We use $Y_{1:t}$ to denote the subsequence of $Y$ ranging from $1$ to $t$ and use $\hat{Y}$ to denote the sequence generated by a model.

\begin{figure}[t!]
    \begin{center}
         \includegraphics[width=\linewidth]{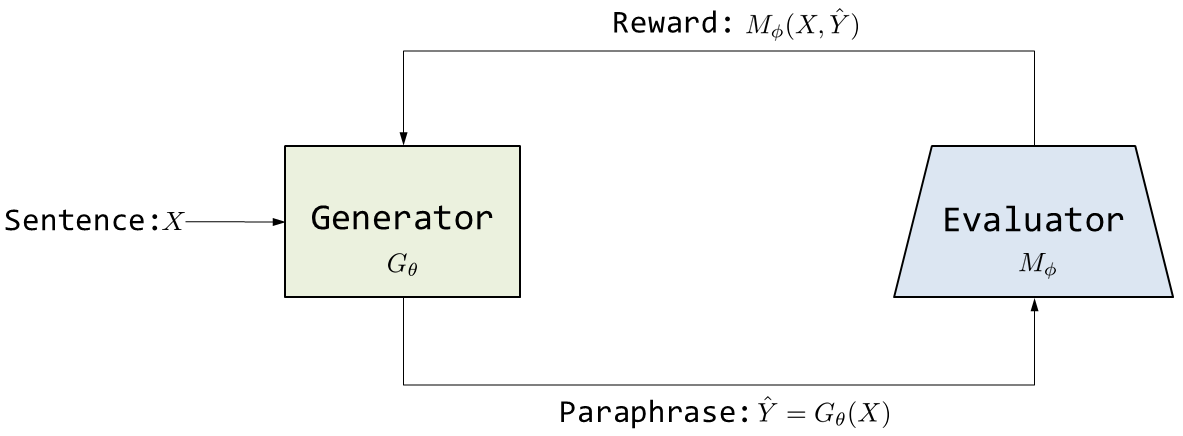}
        \caption{Framework of RbM (Reinforced by Matching).} %%
        \label{fig:pg-rl}
    \end{center}\vspace{-15pt}
    \end{figure}

We propose a framework, which contains a generator and an evaluator, called \textbf{RbM} (Reinforced by Matching). Specifically, for the generator we adopt the Seq2Seq architecture with attention and copy mechanism~\citep{bahdanau2014neural,see2017get}, and for the evaluator we adopt the decomposable attention-based deep matching model~\citep{parikh2016decomposable}. We denote the generator as $G_{\theta}$ and the evaluator as $M_{\phi}$, where $\theta$ and $\phi$ represent their parameters respectively. Figure~\ref{fig:pg-rl} gives an overview of our framework. Basically the generator can generate a paraphrase of a given sentence and the evaluator can judge how semantically similar the two sentences are.

    \subsection{Generator: Seq2Seq Model}\label{sec:seq2seq}
    In this work, paraphrase generation is defined as a sequence-to-sequence (Seq2Seq) learning problem. Given input sentence $X$, the goal is to learn a model $G_{\theta}$ that can generate a sentence $\hat{Y}=G_{\theta}(X)$ as its paraphrase. We choose the \textit{pointer-generator} proposed by ~\citet{see2017get} as the generator. The model is built based on the encoder-decoder framework~\citep{cho2014learning,sutskever2014sequence}, both of which are implemented as recurrent neural networks (RNN). The encoder RNN transforms the input sequence $X$ into a sequence of hidden states $H=[h_1,\ldots,h_S]$. The decoder RNN generates an output sentence $Y$ on the basis of the hidden states. Specifically it predicts the next word at each position by sampling from $\hat{y}_{t} \sim p(y_t|Y_{1:t-1},X)=g(s_{t}, c_{t}, y_{t-1})$, where $s_t$ is the decoder state, $c_t$ is the context vector, $y_{t-1}$ is the previous word, and $g$ is a feed-forward neural network. Attention mechanism~\citep{bahdanau2014neural} is introduced to compute the context vector as the weighted sum of encoder states:
    \[
        c_t=\sum_{i=1}^S\alpha_{ti}h_{i},\ \  \alpha_{ti} = \frac{\exp{\eta(s_{t-1}, h_{i})}}{\sum_{j=1}^S \exp\eta(s_{t-1}, h_{j})},\\
    \]
    where $\alpha_{ti}$ represents the attention weight and $\eta$ is the \textit{attention function}, which is a feed-forward neural network.

    Paraphrasing often needs copying words from the input sentence, for instance, named entities. The pointer-generator model allows either generating words from a vocabulary or copying words from the input sequence. Specifically the probability of generating the next word is given by a mixture model:
    \[
        \begin{aligned}
            p_{\theta}(y_t|Y_{1:t-1}&, X) = q(s_t,c_t,y_{t-1}) g(s_t,c_t,y_{t-1}) \\
            &+(1-q(s_t,c_t,y_{t-1}))\sum\nolimits_{i:y_t=x_i}\alpha_{ti}, \label{eqn:copy}
        \end{aligned}
    \]
    where $q(s_t,c_t,y_{t-1})$ is a binary neural classifier deciding the probability of switching between the generation mode and the copying mode.

    \subsection{Evaluator: Deep Matching Model}\label{sec:matching}

    In this work, paraphrase evaluation (identification) is casted as a problem of learning of sentence matching. The goal is to learn a real-valued function $\MP(X, Y)$ that can represent the matching degree between the two sentences as paraphrases of each other. A variety of learning techniques have been developed for matching sentences, from linear models (e.g., ~\citet{wu2013learning}) to neural network based models (e.g.,~\citet{socher2011dynamic,hu2014convolutional}). We choose a simple yet effective neural network architecture, called the \textit{decomposable-attention} model~\citep{parikh2016decomposable}, as the evaluator. The evaluator can calculate the semantic similarity between two sentences:

    \vspace{-5pt}
    \small \[
        \MP(X, Y) = H(\sum_{i=1}^S G([e(x_i),\bar{x}_i]), \sum_{j=1}^T G([e(y_j), \bar{y}_j])),
    \]\normalsize
    where $e(\cdot)$ denotes a word embedding, $\bar{x}_i$ and $\bar{y}_j$ denote inter-attended vectors, $H$ and $G$ are feed-forward networks. We refer the reader to ~\citet{parikh2016decomposable} for details. In addition, we add \textit{positional encodings} to the word embedding vectors to incorporate the order information of the words, following the idea in~\citet{vaswani2017attention}.

\section{Learning}\label{sec:models}

    This section explains how to learn the generator and evaluator using deep reinforcement learning.

    \subsection{Learning of Generator}\label{sec:seq2seq-rl}
    Given training data $(X,Y)$, the generator $\GT$ is first trained to maximize the conditional log likelihood (negative cross entropy):
        \begin{equation}\label{eqn:seq2seq}
            \mathcal{L}_{\text{Seq2Seq}}(\theta) = \sum\nolimits_{t=1}^T\log p_{\theta}(y_{t}|Y_{1:t-1}, X).
        \end{equation}
    When computing the conditional probability of the next word as above, we choose the previous word $y_{t-1}$ in the ground-truth rather than the word $\hat{y}_{t-1}$ generated by the model. This technique is called \textit{teacher forcing}.

    With teacher forcing, the discrepancy between training and prediction (also referred to as \textit{exposure bias}) can quickly accumulate errors along the generated sequence~\citep{bengio2015scheduled,ranzato2015sequence}. Therefore, the generator $G_{\theta}$ is next fine-tuned using RL, where the reward is given by the evaluator.

    In the RL formulation, generation of the next word represents an \textit{action}, the previous words represent a \textit{state}, and the probability of generation $p_{\theta}(y_{t}|Y_{1:t-1}, X)$ induces a stochastic \textit{policy}. Let $r_t$ denote the \textit{reward} at position $t$. The goal of RL is to find a policy (i.e., a generator) that maximizes the expected cumulative reward:
        \begin{equation}\label{eqn:rl}
            \mathcal{L}_{RL}(\theta) = \mathbb{E}_{p_{\theta}(\hat{Y}|X)}\sum_{t=1}^{T}r_t(X, \hat{Y}_{1:t}).
            %\max_{\theta} \mathbb{E}_{p_{\theta}(Y_{1:T}|X)}\sum_{t=1}^{t=T}r_t(X, Y_{1:t}).
        \end{equation}

    We define a positive reward at the end of sequence ($r_T=R$) and a zero reward at the other positions. The reward $R$ is given by the evaluator $M_{\phi}$. In particular, when a pair of input sentence $X$ and generated paraphrase $\hat{Y}=\GT(X)$ is given, the reward is calculated by the evaluator $R=M_{\phi}(X,\hat{Y})$.

    We can then learn the optimal policy by employing policy gradient. According to the policy gradient theorem~\citep{williams1992simple,sutton2000policy}, the gradient of the expected cumulative reward can be calculated by
    \begin{equation}\label{eqn:policy-gradient}
        \nabla_{\theta} \mathcal{L}_{RL}(\theta) = \sum_{t=1}^T[\nabla_{\theta}\log p_{\theta}(\hat{y}_{t}|\hat{Y}_{1:t-1}, X)]r_t.
    \end{equation}
    The generator can thus be learned  with stochastic gradient descent methods such as Adam~\citep{kingma2014adam}.

    \subsection{Learning of Evaluator}\label{sec:seq2seq-matching-rl}

    The evaluator works as the reward function in RL of the generator and thus is essential for the task. We propose two methods for learning the evaluator in different settings. When there are both positive and negative examples of paraphrases, the evaluator is trained by \textit{supervised learning} (SL). When only positive examples are available (usually the same data as the training data of the generator), the evaluator is trained by \textit{inverse reinforcement learning} (IRL).
    \vspace{-5pt}
    \subsubsection*{Supervised Learning}
    Given a set of positive and negative examples (paraphrase pairs), we conduct supervised learning of the evaluator with the pointwise cross entropy loss:
        \begin{equation}\label{eqn:match-pointwise}
            \mathcal{J}_{\text{SL}}(\phi) = -\log\MP(X, Y)-\log(1-\MP(X, Y^{-})),
        \end{equation}
    where $Y^{-}$ represents a sentence that is not a paraphrase of $X$. The evaluator $\MP$ here is defined as a classifier, trained to distinguish negative example $(X, Y^{-})$ from positive example $(X, Y)$.

    We call this method \textbf{RbM-SL} (Reinforced by Matching with Supervised Learning). The evaluator $\MP$ trained by supervised learning can make a judgement on whether two sentences are paraphrases of each other. With a well-trained evaluator $\MP$, we further train the generator $G_{\theta}$ by reinforcement learning using $\MP$ as a reward function. Figure~\ref{fig:lp-sl} shows the learning process of RbM-SL. The detailed training procedure is shown in Algorithm 1 in Appendix A.

    \begin{figure*}[htp]
        \centering
            \subfloat[RbM-SL]{%
            \includegraphics[width=0.45\textwidth]{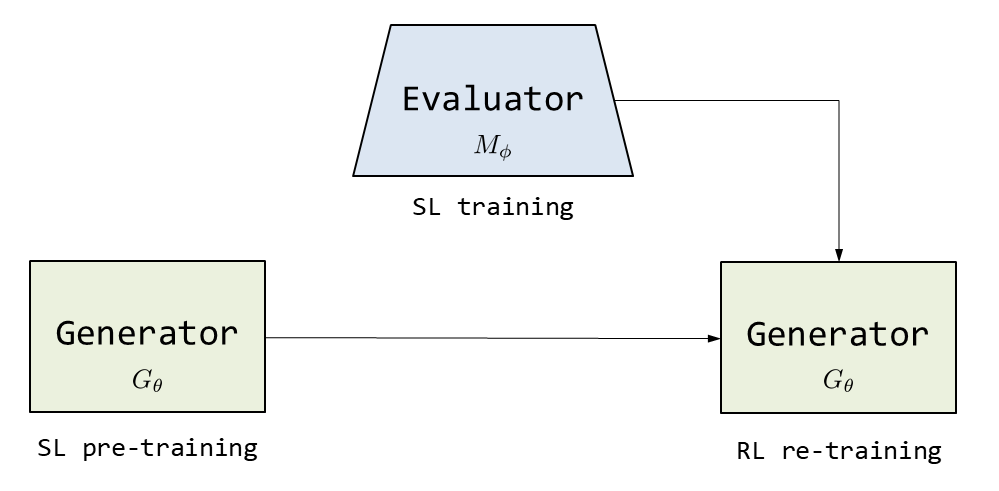}%
            \label{fig:lp-sl}%
        }%
        \hfill%
            \subfloat[RbM-IRL]{%
        \includegraphics[width=0.5\textwidth]{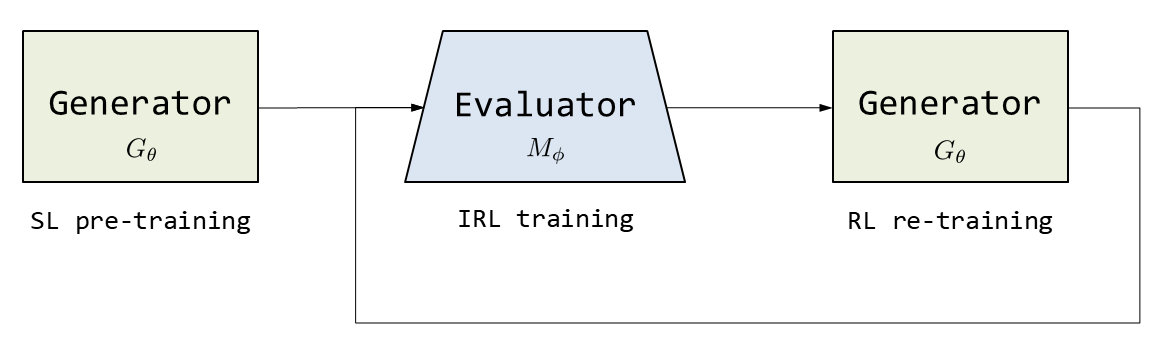}%
        \label{fig:lp-irl}%
        }%
        \caption{Learning Process of RbM models: (a) RbM-SL, (b) RbM-IRL.} %%
    \end{figure*}

    \subsubsection*{Inverse Reinforcement Learning}
    Inverse reinforcement learning (IRL) is a sub-problem of reinforcement learning (RL), about learning of a reward function given \textit{expert demonstrations}, which are sequences of states and actions from an expert (optimal) policy. More specifically, the goal is to find an optimal reward function $R^*$ with which the expert policy $p_{\theta^*}(Y|X)$ really becomes optimal among all possible policies, i.e.,
        \[
          \mathbb{E}_{p_{\theta^*}(Y|X)} R^*(Y) \geq \mathbb{E}_{p_{\theta}(\hat{Y}|X)} R^*(\hat{Y}), \ \ \forall \theta.
        \]

    In the current problem setting, the problem becomes learning of an optimal reward function (evaluator) given a number of paraphrase pairs given by human experts (expert demonstrations).

    To learn an optimal reward (matching) function is challenging, because the expert demonstrations might not be optimal and the reward function might not be rigorously defined. To deal with the problem, we employ the maximum margin formulation of IRL inspired by~\citet{ratliff2006maximum}.

   The maximum margin approach ensures the learned reward function has the following two desirable properties in the paraphrase generation task: (a) given the same input sentence, a reference from humans should have a higher reward than the ones generated by the model; (b) the margins between the rewards should become smaller when the paraphrases generated by the model get closer to a reference given by humans. We thus specifically consider the following optimization problem for learning of the evaluator:
        \begin{equation}
            \mathcal{J}_{\text{IRL}}(\phi) = \max (0, 1  - \zeta + M_{\phi}(X, \hat{Y}) - M_{\phi}(X, Y)), \label{eqn:max-margin-irl}
        \end{equation}
    where $\zeta$ is a slack variable to measure the agreement between $\hat{Y}$ and $Y$. In practice we set $\zeta = \text{ROUGE-L}(\hat{Y}, Y)$. Different from RbM-SL, the evaluator $\MP$ here is defined as a ranking model that assigns higher rewards to more plausible paraphrases.

    Once the reward function (evaluator) is learned, it is then used to improve the policy function (generator) through policy gradient.
    In fact, the generator $\GT$ and the evaluator $\MP$ are trained alternatively. We call this method \textbf{RbM-IRL} (Reinforced by Matching with Inverse Reinforcement Learning).
    Figure~\ref{fig:lp-irl} shows the learning process of RbM-IRL. The detailed training procedure is shown in Algorithm 2 in Appendix A.

    We can formalize the whole learning procedure as the following optimization problem:
        \begin{equation}\label{eqn:irl}
          %\min_{\phi} \max_{\theta} [\mathbb{E}_{p_{\theta}(\hat{Y}|X)} M_{\phi}(X, \hat{Y}) - \mathbb{E}_{p_{\theta^*}(Y|X)} M_{\phi}(X, Y)]
          \min_{\phi} \max_{\theta} \mathbb{E}_{p_{\theta}(\hat{Y}|X)} \mathcal{J}_{\text{IRL}}(\phi).
        \end{equation}

    RbM-IRL can make effective use of sequences generated by the generator for training of the evaluator. As the generated sentences become closer to the ground-truth, the evaluator also becomes more discriminative in identifying paraphrases.

    It should be also noted that for both RbM-SL and RbM-IRL, once the evaluator is learned, the reinforcement learning of the generator only needs \textit{non-parallel sentences} as input. This makes it possible to further train the generator and enhance the generalization ability of the generator.

    \subsection{Training Techniques}\label{sec:training-techniques}
    \subsubsection*{Reward Shaping}\label{sec:reward-shaping}
    In the original RL of the generator, only a positive reward $R$ is given at the end of sentence. This provides sparse supervision signals and can make the model greatly degenerate. Inspired by the idea of reward shaping \citep{ng1999policy,bahdanau2016actor}, we estimate the intermediate cumulative reward (value function) for each position, that is
    \[
        Q_t=\mathbb{E}_{p_{\theta}(Y_{t+1:T}|\hat{Y}_{1:t},X)} R(X, [\hat{Y}_{1:t},Y_{t+1:T}]),
    \]
    by Monte-Carlo simulation, in the same way as in~\citet{yu2017seqgan}:
    \begin{equation} \label{eqn:match-reward}
        Q_t =
            \begin{cases}
                \frac{1}{N}\sum_{n=1}^{n=N}\MP(X, [\hat{Y}_{1:t}, \widehat{Y}_{t+1:T}^n]), & t < T\\
                \MP(X, \hat{Y}), & t = T,
            \end{cases}
    \end{equation}
    where $N$ is the sample size and $\widehat{Y}_{t+1:T}^n \sim p_{\theta}(Y_{t+1:T}|\hat{Y}_{1:t},X)$ denotes simulated sub-sequences randomly sampled starting from the $(t+1)$-th word. During training of the generator, the reward $r_t$ in policy gradient \eqref{eqn:policy-gradient} is replaced by $Q_t$ estimated in \eqref{eqn:match-reward}.

    \subsubsection*{Reward Rescaling}\label{sec:reward-rescale}
    In practice, RL algorithms often suffer from instability in training. A common approach to reduce the variance is to subtract a baseline reward from the value function. For instance, a simple baseline can be a moving average of historical rewards. %More advanced methods like the actor-critic algorithms employ another trainable model (the critic) to approximate the expected reward.
    While in RbM-IRL, the evaluator keeps updating during training. Thus, keeping track of a baseline reward is unstable and inefficient. Inspired by~\citet{guo2017long}, we propose an efficient reward rescaling method based on ranking. For a batch of $D$ generated paraphrases $\{\hat{Y}^{d}\}_{d=1}^D$, each associated with a reward $R^{d}=\MP(X^d, \hat{Y}^d)$, we rescale the rewards by
        \begin{equation}\label{eqn:rescale-1}
            \bar{R}^d = \sigma(\delta_1 \cdot (0.5-\frac{\text{rank}(d)}{D})) - 0.5,
        \end{equation}
    where $\sigma(\mathord{\cdot})$ is the sigmoid function, $\text{rank}(d)$ is the rank of $R^{d}$ in $\{R^{1},...,R^{D}\}$, and $\delta_1$ is a scalar controlling the variance of rewards. A similar strategy is applied into estimation of in-sequence value function for each word, and the final rescaled value function is
        \begin{equation}\label{eqn:rescale-2}
            \bar{Q}^d_t = \sigma(\delta_2 \cdot (0.5-\frac{\text{rank}(t)}{T})) - 0.5 + \bar{R}^d,
        \end{equation}
    where $\text{rank}(t)$ is the rank of $Q^{d}_t$ in $\{Q^{d}_1,...,Q^{d}_T\}$.

    Reward rescaling has two advantages. First, the mean and variance of $\bar{Q}_t^d$ are controlled and hence they make the policy gradient more stable, even with a varying reward function. Second, when the evaluator $\MP$ is trained with the ranking loss as in RbM-IRL, it is better to inform which paraphrase is better, rather than to provide a scalar reward in a range. In our experiment, we find that this method can bring substantial gains for RbM-SL and RbM-IRL, but not for RL with ROUGE as reward.

    \subsubsection*{Curriculum Learning}
    RbM-IRL may not achieve its best performance if all of the training instances are included in training at the beginning. We employ a curriculum learning strategy~\citep{bengio2009curriculum} for it. During the training of the evaluator $\MP$, each example $k$ is associated with a weight $w^k$, i.e.
    \vspace{-5pt}
        \begin{align}
            \mathcal{J}^k_{\text{IRL-CL}}(\phi) =  w^k \max (0, & 1 - \zeta^k + \nonumber\\
             M_{\phi}(X^k, \hat{Y}^k) & - M_{\phi}(X^k, Y^k) ) \label{eqn:cl-1}
        \end{align}

    In curriculum learning, $w^k$ is determined by the difficulty of the example. At the beginning, the training procedure concentrates on relatively simple examples, and gradually puts more weights on difficult ones. In our case, we use the edit distance $\mathcal{E}(\mathord{X}, \mathord{Y})$ between $X$ and $Y$ as the measure of difficulty for paraphrasing. Specifically, $w^k$ is determined by $w^k \sim \text{Binomial}(p^k, 1)$, and $p^k = \sigma(\delta_3 \cdot (0.5-\frac{\text{rank}(\mathcal{E}(X^k, Y^k))}{K}))$, where $K$ denotes the batch size for training the evaluator. For $\delta_3$, we start with a relatively high value and gradually decrease it. In the end each example will be sampled with a probability around $0.5$. In this manner, the evaluator first learns to identify paraphrases with small modifications on the input sentences (e.g. ``\textit{what 's}'' and ``\textit{what is}''). Along with training it gradually learns to handle more complicated paraphrases (e.g. ``\textit{how can I}'' and ``\textit{what is the best way to}'').

\section{Experiment}\label{sec:exp}

    \subsection{Baselines and Evaluation Measures}
    To compare our methods (RbM-SL and RbM-IRL) with existing neural network based methods, we choose five baseline models: the attentive Seq2Seq model~\citep{bahdanau2014neural}, the stacked Residual LSTM networks~\citep{prakash2016neural}, the variational auto-encoder (VAE-SVG-eq)~\citep{gupta2017deep}~\footnote{We directly present the results reported in ~\citet{gupta2017deep} on the same dateset and settings.}, the pointer-generator~\citep{see2017get}, and the reinforced pointer-generator with ROUGE-2 as reward (RL-ROUGE)~\citep{ranzato2015sequence}.

    We conduct both automatic and manual evaluation on the models. For the automatic evaluation, we adopt four evaluation measures: ROUGE-1, ROUGE-2~\citep{lin2004rouge}, BLEU~\citep{papineni2002bleu} (up to at most bi-grams) and METEOR~\citep{lavie2007meteor}. As pointed out, ideally it would be better not to merely use a lexical measure like ROUGE or BLEU for evaluation of paraphrasing. We choose to use them for reproducibility of our experimental results by others. For the manual evaluation, we conduct evaluation on the generated paraphrases in terms of relevance and fluency.

    \subsection{Datasets}
    We evaluate our methods with the \textbf{Quora} question pair dataset~\footnote{\url{https://www.kaggle.com/c/quora-question-pairs}} and \textbf{Twitter} URL paraphrasing corpus~\citep{lan2017continuously}. Both datasets contain positive and negative examples of paraphrases so that we can evaluate the RbM-SL and RbM-IRL methods. We randomly split the Quora dataset in two different ways obtaining two experimental settings: Quora-\Rmnum{1} and Quora-\Rmnum{2}. In Quora-\Rmnum{1}, we partition the dataset by question pairs, while in Quora-\Rmnum{2}, we partition by question ids such that there is no shared question between the training and test/validation datasets. In addition, we sample a smaller training set in Quora-\Rmnum{2} to make the task more challenging.  Twitter URL paraphrasing corpus contains two subsets, one is labeled by human annotators while the other is labeled automatically by algorithm. We sample the test and validation set from the labeled subset, while using the remaining pairs as training set. For RbM-SL, we use the labeled subset to train the evaluator $\MP$. Compared to Quora-\Rmnum{1}, it is more difficult to achieve a high performance with Quora-\Rmnum{2}. The Twitter corpus is even more challenging since the data contains more noise. The basic statistics of datasets are shown in Table \ref{tab:data}.
        \begin{table}[h!]
            \centering
            \caption{Statistics of datasets.}\label{tab:data}
            \vspace{-5pt}
            \resizebox{\linewidth}{!}{
            \begin{tabular}{lcccccccc}
                \toprule
                & \multicolumn{3}{c}{Generator} & \multicolumn{2}{c}{Evaluator (RbM-SL)}\\
                \cmidrule(lr){2-4}\cmidrule(lr){5-6}
                Dataset & \#Train & \#Test & \#Validation & \#Positive & \#Negative \\
                \midrule
                Quora-\Rmnum{1} & 100K & 30K & 3K & 100K & 160K \\
                Quora-\Rmnum{2} & 50K & 30K & 3K & 50K & 160K\\
                Twitter & 110K & 5K & 1K & 10K & 40K\\
                \bottomrule
            \end{tabular}
            }
        \end{table}

        \begin{table*}[t!]
            \centering
            \caption{Performances on Quora datasets.}\label{tab:quora}\vspace{-10pt}
            \resizebox{0.9\linewidth}{!}{
            \begin{tabular}{lccccccccc}
                \toprule
                & \multicolumn{4}{c}{Quora-\Rmnum{1}} & & \multicolumn{4}{c}{Quora-\Rmnum{2}}\\
                \cmidrule(lr){2-5}\cmidrule(lr){7-10}
                Models & ROUGE-1 & ROUGE-2 & BLEU & METEOR & & ROUGE-1 & ROUGE-2 & BLEU & METEOR\\
                \midrule
                Seq2Seq & 58.77 & 31.47 & 36.55 & 26.28 & & 47.22 & 20.72 & 26.06 & 20.35 \\
                Residual LSTM & 59.21 & 32.43  & 37.38 &  28.17 & & 48.55 & 22.48 & 27.32 & 22.37 \\
                VAE-SVG-eq & - & -  & - &  25.50 & & - & - & - & 22.20 \\
                Pointer-generator & 61.96 & 36.07 & 40.55 & 30.21 & & 51.98 & 25.16 & 30.01 & 24.31\\
                RL-ROUGE & 63.35 & 37.33 & 41.83 & 30.96 & & 54.50 & 27.50 & 32.54 & 25.67\\
                \midrule
                RbM-SL (ours) & \textbf{64.39} & \textbf{38.11} & \textbf{43.54} & \textbf{32.84} & & \textbf{57.34} & \textbf{31.09} & \textbf{35.81} & \textbf{28.12} \\
                RbM-IRL (ours) & 64.02 & 37.72 & 43.09  & 31.97 & & 56.86 & 29.90 & 34.79 & 26.67\\
                \bottomrule
            \end{tabular}
            }
        \end{table*}

        \begin{table*}[t!]
            \parbox[t]{0.5\linewidth}{
                \centering
                \caption{Performances on Twitter corpus.}\label{tab:twitter}\vspace{-10pt}
                \resizebox{\linewidth}{!}{
                \begin{tabular}{lcccc}
                    \toprule
                     & \multicolumn{4}{c}{Twitter}\\
                     \cmidrule(lr){2-5}
                    Models & ROUGE-1 & ROUGE-2 & BLEU & METEOR\\
                    \midrule
                    Seq2Seq & 30.43 & 14.61 & 30.54 & 12.80  \\
                    Residual LSTM & 32.50 & 16.86 & 33.90 & 13.65 \\
                    Pointer-generator & 38.31 & 21.22 & 40.37 & 17.62 \\
                    RL-ROUGE & 40.16 & 22.99 & 42.73 & 18.89 \\
                    \midrule
                    RbM-SL (ours) & 41.87 & 24.23 & 44.67 & 19.97 \\
                    RbM-IRL  (ours) & \textbf{42.15} & \textbf{24.73} & \textbf{45.74} & \textbf{20.18} \\
                    \bottomrule
                \end{tabular}}
            }
            \hfill
            \parbox[t]{0.5\linewidth}{
                \centering
                \caption{Human evaluation on Quora datasets.}\label{tab:quora-human}\vspace{-10pt}
                \resizebox{\linewidth}{!}{
                \begin{tabular}{lcccc}
                    \toprule
                    & \multicolumn{2}{c}{Quora-\Rmnum{1}} & \multicolumn{2}{c}{Quora-\Rmnum{2}}\\
                    \cmidrule(lr){2-3}\cmidrule(lr){4-5}
                    Models & Relevance & Fluency & Relevance & Fluency \\
                    \midrule
                    Pointer-generator & 3.23 & 4.55 & 2.34 & 2.96\\
                    RL-ROUGE & 3.56 & 4.61 & 2.58 & 3.14\\
                    \midrule
                    RbM-SL (ours) & \textbf{4.08} & 4.67 & \textbf{3.20} & 3.48\\
                    RbM-IRL (ours) & 4.07 & \textbf{4.69} & 2.80 & \textbf{3.53}\\
                    \midrule \midrule
                    Reference & 4.69 & 4.95 & 4.68 & 4.90\\
                    \bottomrule
                \end{tabular}
                }
            }
            \end{table*}

    \subsection{Implementation Details}
    \noindent \textbf{Generator} We maintain a fixed-size vocabulary of 5K shared by the words in input and output, and truncate all the sentences longer than 20 words. The model architecture, word embedding size and LSTM cell size are as the same as reported in~\citet{see2017get}.
    We use Adadgrad optimizer~\citep{duchi2011adaptive} in the supervised pre-training and Adam optimizer in the reinforcement learning, with the batch size of 80. We also fine-tune the Seq2Seq baseline models with Adam optimizer for a fair comparison. In supervised pre-training, we set the learning rate as 0.1 and initial accumulator as 0.1. The maximum norm of gradient is set as 2. During the RL training, the learning rate decreases to 1e-5 and the size of Monte-Carlo sample is 4. To make the training more stable, we use the ground-truth with reward of 0.1.

    \noindent \textbf{Evaluator} We use the pretrained GoogleNews 300-dimension word vectors~\footnote{\url{https://code.google.com/archive/p/word2vec/}} in Quora dataset and 200-dimension GloVe word vectors~\footnote{\url{https://nlp.stanford.edu/projects/glove/}} in Twitter corpus. Other model settings are the same as in~\citet{parikh2016decomposable}. For evaluator in RbM-SL we set the learning rate as 0.05 and the batch size as 32.  For the evaluator of $\MP$ in RbM-IRL, the learning rate decreases to 1e-2, and we use the batch size of 80.

    We use the technique of reward rescaling as mentioned in section~\ref{sec:reward-rescale} in training RbM-SL and RbM-IRL. In RbM-SL, we set $\delta_{1}$ as 12 and $\delta_{2}$ as 1. In RbM-IRL, we keep $\delta_{2}$ as 1 all the time and decrease $\delta_{1}$ from 12 to 3 and $\delta_{3}$ from 15 to 8 during curriculum learning. In ROUGE-RL, we take the exponential moving average of historical rewards as baseline reward to stabilize the training:
    \[
        b_{m} = \lambda \overline{Q}_{m-1} + (1-\lambda) b_{m-1},\ b_1=0
    \]
    where $b_{m}$ is the baseline $b$ at iteration $m$, $\overline{Q}$ is the mean value of reward, and we set $\lambda$ as 0.1 by grid search.

    \subsection{Results and Analysis}
    \noindent \textbf{Automatic evaluation}\  Table~\ref{tab:quora} shows the performances of the models on Quora datasets. In both settings, we find that the proposed RbM-SL and RbM-IRL models outperform the baseline models in terms of all the evaluation measures. Particularly in Quora-\Rmnum{2}, RbM-SL and RbM-IRL make significant improvements over the baselines, which demonstrates their higher ability in learning for paraphrase generation. On Quora dataset, RbM-SL is constantly better than RbM-IRL for all the automatic measures, which is reasonable because RbM-SL makes use of additional labeled data to train the evaluator. Quora datasets contains a large number of high-quality non-paraphrases, i.e., they are literally similar but semantically different, for instance ``\textit{are analogue clocks better than digital}'' and ``\textit{is analogue better than digital}''. Trained with the data, the evaluator tends to become more capable in paraphrase identification. With additional evaluation on Quora data, the evaluator used in RbM-SL can achieve an accuracy of 87\% on identifying positive and negative pairs of paraphrases.

    Table~\ref{tab:twitter} shows the performances on the Twitter corpus. Our models again outperform the baselines in terms of all the evaluation measures. Note that RbM-IRL performs better than RbM-SL in this case. The reason might be that the evaluator of RbM-SL might not be effectively trained with the relatively small dataset, while RbM-IRL can leverage its advantage in learning of the evaluator with less data.

    In our experiments, we find that the training techniques proposed in section \ref{sec:training-techniques} are all necessary and effective. Reward shaping is by default employed by all the RL based models. Reward rescaling works particularly well for the RbM models, where the reward functions are learned from data. Without reward rescaling, RbM-SL can still outperform the baselines but with smaller margins. For RbM-IRL, curriculum learning is necessary for its best performance. Without curriculum learning, RbM-IRL only has comparable performance with ROUGE-RL.

    \begin{figure*}[t!]
      \centering
      \caption{Examples of the generated paraphrases by different models on Quora-\Rmnum{2}.}\label{fig:examples}\vspace{-10pt}
      \includegraphics[width=0.9\textwidth]{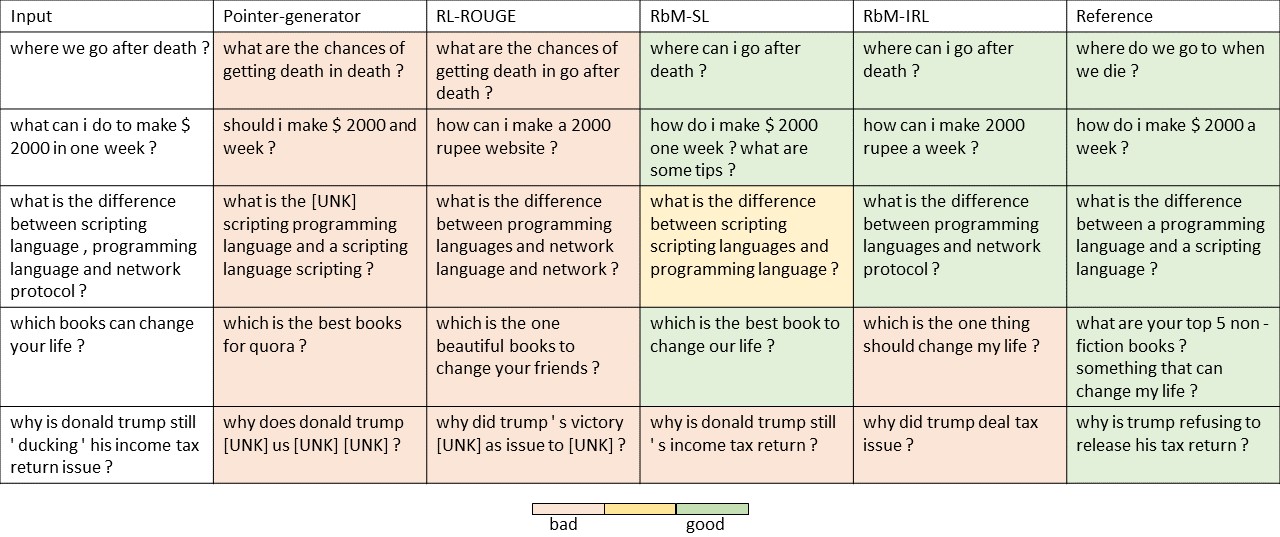}\vspace{-10pt}
    \end{figure*}

    \noindent \textbf{Human evaluation}\  We randomly select 300 sentences from the test data as input and generate paraphrases using different models. The pairs of paraphrases are then aggregated and partitioned into seven random buckets for seven human assessors to evaluate. The assessors are asked to rate each sentence pair according to the following two criteria: \textit{relevance} (the paraphrase sentence is semantically close to the original sentence) and \textit{fluency} (the paraphrase sentence is fluent as a natural language sentence, and the grammar is correct). Hence each assessor gives two scores to each paraphrase, both ranging from 1 to 5. To reduce the evaluation variance, there is a detailed evaluation guideline for the assessors in Appendix B. Each paraphrase is rated by two assessors, and then averaged as the final judgement. The agreement between assessors is moderate (kappa=0.44).

    Table~\ref{tab:quora-human} demonstrates the average ratings for each model, including the ground-truth references. Our models of RbM-SL and RbM-IRL get better scores in terms of relevance and fluency than the baseline models, and their differences are statistically significant (paired \textit{t}-test, \textit{p}-value $< 0.01$). We note that in human evaluation, RbM-SL achieves the best relevance score while RbM-IRL achieves the best fluency score.

    \noindent \textbf{Case study}\  Figure~\ref{fig:examples} gives some examples of generated paraphrases by the models on Quora-\Rmnum{2} for illustration. The first and second examples show the superior performances of RbM-SL and RbM-IRL over the other models. In the third example, both RbM-SL and RbM-IRL capture accurate paraphrasing patterns, while the other models wrongly segment and copy words from the input sentence. Compared to RbM-SL with an error of repeating the word \textit{scripting}, RbM-IRL generates a more fluent paraphrase. The reason is that the evaluator in RbM-IRL is more capable of measuring the fluency of a sentence. In the fourth example, RL-ROUGE generates a totally non-sense sentence, and pointer-generator and RbM-IRL just cover half of the content of the original sentence, while RbM-SL successfully rephrases and preserves all the meaning. All of the models fail in the last example, because the word \textit{ducking} is a rare word that never appears in the training data. Pointer-generator and RL-ROUGE generate totally irrelevant words such as \textit{UNK} token or \textit{victory}, while RbM-SL and RbM-IRL still generate topic-relevant words.
    \vspace{-5pt}

\section{Related Work}\label{seq:relate}

    %\textbf{Sequence-to-sequence learning} (Seq2Seq) is first proposed in~\citet{sutskever2014sequence,cho2014learning}. The attention mechanism is introduced into Seq2Seq to solve the alignment problem in machine translation~\citet{bahdanau2014neural}. The copy mechanism is added to Seq2Seq to tackle the out of vocabulary problem~\citep{gu2016incorporating, see2017get}. \citet{ranzato2015sequence,bahdanau2016actor} formalize the learning of Seq2Seq model as a reinforcement learning problem, in which the reward function is manually defined and policy gradient is utilized. In our work, we basically adopt the Seq2Seq model with attention and copy mechanism. We also cast the problem as reinforcement learning, but the reward function used is defined by another trainable deep model, i.e., the evaluator.

    \textbf{Neural paraphrase generation} recently draws attention in different application scenarios. The task is often formalized as a sequence-to-sequence (Seq2Seq) learning problem. \citet{prakash2016neural} employ a stacked residual LSTM network in the Seq2Seq model to enlarge the model capacity. ~\citet{cao2017joint} utilize an additional vocabulary to restrict word candidates during generation. ~\citet{gupta2017deep} use a variational auto-encoder framework to generate more diverse paraphrases. ~\citet{ma2018word} utilize an attention layer instead of a linear mapping in the decoder to pick up word candidates. ~\citet{iyyer2018adversarial} harness syntactic information for controllable paraphrase generation. \citet{zhang2017sentence} tackle a similar task of sentence simplification withe Seq2Seq model coupled with deep reinforcement learning, in which the reward function is manually defined for the task. Similar to these works, we also pre-train the paraphrase generator within the Seq2Seq framework. The main difference lies in that we use another trainable neural network, referred to as evaluator, to guide the training of the generator through reinforcement learning.

    There is also work on paraphrasing generation in different settings. For example, \citet{mallinson2017paraphrasing} leverage bilingual data to produce paraphrases by pivoting over a shared translation in another language. \citet{wieting2017learning,wieting2018paranmt} use neural machine translation to generate paraphrases via back-translation of bilingual sentence pairs. \citet{buck2017ask} and \citet{dong2017learning} tackle the problem of QA-specific paraphrasing with the guidance from an external QA system and an associated evaluation metric.

    \textbf{Inverse reinforcement learning} (IRL) aims to learn a reward function from expert demonstrations. \citet{abbeel2004apprenticeship} propose \textit{apprenticeship learning}, which uses a feature based linear reward function and learns to match feature expectations. \citet{ratliff2006maximum} cast the problem as structured maximum margin prediction. \citet{ziebart2008maximum} propose max entropy IRL in order to solve the problem of expert suboptimality. Recent work involving deep learning in IRL includes \citet{finn2016guided} and \citet{ho2016model}. There does not seem to be much work on IRL for NLP. In ~\citet{neu2009training}, parsing is formalized as a feature expectation matching problem. ~\citet{wang2018AREL} apply adversarial inverse reinforcement learning in visual story telling. To the best of our knowledge, our work is the first that applies deep IRL into a Seq2Seq task.

    \textbf{Generative Adversarial Networks} (GAN)~\citep{goodfellow2014generative} is a family of unsupervised generative models. GAN contains a generator and a discriminator, respectively for generating examples from random noises and distinguishing generated examples from real examples, and they are trained in an adversarial way. There are applications of GAN on NLP, such as text generation~\citep{yu2017seqgan,guo2017long} and dialogue generation~\citep{li2017adversarial}. RankGAN \citep{lin2017adversarial} is the one most similar to RbM-IRL that employs a ranking model as the discriminator. However, RankGAN works for text generation rather than sequence-to-sequence learning, and training of generator in RankGAN relies on parallel data while the training of RbM-IRL can use non-parallel data.

    There are connections between GAN and IRL as pointed by~\citet{finn2016connection, ho2016generative}. However, there are significant differences between GAN and our RbM-IRL model. GAN employs the discriminator to distinguish generated examples from real examples, while RbM-IRL employs the evaluator as a reward function in RL. The generator in GAN is trained to maximize the loss of the discriminator in an adversarial way, while the generator in RbM-IRL is trained to maximize the expected cumulative reward from the evaluator.

\vspace{-5pt}
\section{Conclusion}
\vspace{-5pt}
    In this paper, we have proposed a novel deep reinforcement learning approach to paraphrase generation, with a new framework consisting of a generator and an evaluator, modeled as sequence-to-sequence learning model and deep matching model respectively. The generator, which is for paraphrase generation, is first trained via sequence-to-sequence learning. The evaluator, which is for paraphrase identification, is then trained via supervised learning or inverse reinforcement learning in different settings. With a well-trained evaluator, the generator is further fine-tuned by reinforcement learning to produce more accurate paraphrases. The experiment results demonstrate that the proposed method can significantly improve the quality of paraphrase generation upon the baseline methods. In the future, we plan to apply the framework and training techniques into other tasks, such as machine translation and dialogue.

\section*{Acknowledgments}
\vspace{-5pt}
This work is supported by China National 973 Program 2014CB340301.

\bibliography{reference}
\bibliographystyle{acl_natbib}

\newpage
\appendix
\section{Algorithms of RbM-SL and RbM-IRL}
    \begin{algorithm}[h!tb]
    \caption{Training Procedure of RbM-SL}\label{alg:rbm}
    \SetKwInOut{Input}{Input}
    \SetKwInOut{Output}{Output}
    \SetKw{Return}{Return}

    \Input{A corpus of paraphrase pairs $\{(X, Y)\}$, a corpus of non-paraphrase pairs $\{(X, Y^{-})\}$, a corpus of (non-parallel) sentences $\{X\}$.}
    \Output{Generator $G_{\theta'}$}
    Train the evaluator $\MP$ with $\{(X, Y)\}$ and $\{(X, Y^{-})\}$\;
    Pre-train the generator $\GT$ with $\{(X, Y)\}$\;
    Init $G_{\theta'}:=\GT$\;
    \While{not converge}{
        Sample a sentence $X=[x_1,\ldots,x_S]$ from the paraphrase corpus or the non-parallel corpus\;
        Generate a sentence $\hat{Y}=[\hat{y}_1,\ldots,\hat{y}_T]$ according to $G_{\theta'}$\ given input $X$\;
        Set the gradient $g_{\theta'}=0$\;
        \For{$t = 1$ \KwTo $T$}{\
            Run $N$ Monte Carlo simulations: $\{\widehat{Y}_{t+1:T}^1,...\widehat{Y}_{t+1:T}^N\} \sim p_{\theta'}(Y_{t+1:T}|\hat{Y}_{1:t},X)$\;
            Compute the value function by \[
            Q_t =
            \begin{cases}
                \frac{1}{N}\sum_{n=1}^{N} \MP(X,[\hat{Y}_{1:t},\widehat{Y}_{t+1:T}^n]), &\ t < T\\
                \MP(X,\hat{Y}), &\ t=T.
            \end{cases}\]
            Rescale the reward to $\bar{Q}_t$ by \eqref{eqn:rescale-1}\;
            Accumulate $\theta'$-gradient: $ g_{\theta'} := g_{\theta'} + \nabla_{\theta}\log p_{\theta'}(\hat{y}_{t}|\hat{Y}_{1:t-1}, X) \bar{Q}_t$
        }
        Update $G_{\theta'}$ using the gradient $g_{\theta'}$ with learning rate $\gamma_{G}$: $ G_{\theta'} := G_{\theta'} + \gamma_{G} g_{\theta'} $
    }
    \Return{$G_{\theta'}$}
    \end{algorithm}
\newpage
    \begin{algorithm}[h!tb]
        \caption{Training Procedure of RbM-IRL}\label{alg:irbm}
        \SetKwInOut{Input}{Input}
        \SetKwInOut{Output}{Output}
        \SetKw{Return}{Return}

        \Input{A corpus of paraphrase pairs $\{(X, Y)\}$, a corpus of (non-parallel) sentences $\{X\}$.}
        \Output{Generator $G_{\theta'}$, evaluator $M_{\phi'}$}
        %Train the matching network $\MP$ with $\{(X, Y)\}$ and $\{(X, Y^{-})\}$\;
        Pre-train the generator $\GT$ with $\{(X, Y)\}$\;
        Init $G_{\theta'}:=\GT$\ and $M_{\phi'}$\;
        \While{not converge}{
            \While{not converge}{\
                Sample a sentence $X=[x_1,\ldots,x_S]$ from the paraphrase corpus\;
                Generate a sentence $\hat{Y}=[\hat{y}_1,\ldots,\hat{y}_T]$ according to $G_{\theta'}$\ given input $X$\;
                Calculate $\phi'$-gradient: $g_{\phi'}:= \nabla_{\phi}\mathcal{J}_{\text{IRL-CL}}(\phi)$\;
                Update $M_{\phi'}$ using the gradient $g_{\phi'}$ with learning rate $\gamma_{M}$: $M_{\phi'} := M_{\phi'} - \gamma_{M}g_{\phi'}$
            }
            Train $G_{\theta'}$ with $M_{\phi'}$ as in Algorithm \ref{alg:rbm}\;
        }
        \Return{$G_{\theta'}$, $M_{\phi'}$}
    \end{algorithm}
\clearpage

%\onecolumn
\section{Human Evaluation Guideline}
Please judge the paraphrases from the following two criteria:
\begin{itemize}[leftmargin=18pt,topsep=1pt]
  \setlength \itemsep{-0.1em}
  \item[(1)] {\bf Grammar and Fluency}: the paraphrase is acceptable as natural language text, and the grammar is correct;
  \item[(2)] {\bf Coherent and Consistent}: please view from the perspective of the original poster, to what extent the answer of paraphrase is
      helpful for you with respect to the original question. Specifically, you can consider following aspects:
      \begin{itemize}
        \item Relatedness: it should be topically relevant to the original question.
        \item Type of question: the type of the original question remains the same in paraphrase.
        \item Informative: no information loss in paraphrase.
      \end{itemize}
\end{itemize}
For each paraphrase, give two separate score ranking from 1 to 5. The meaning of specific score is as following:
    \begin{itemize}
      \item Grammar and Fluency
        \begin{itemize}
          \item 5: Without any grammatical error;
          \item 4: Fluent and has one minor grammatical error that does not affect understanding, e.g. \textit{what is the best ways to learn programming};
          \item 3: Basically fluent and has two or more minor grammatical errors or one serious grammatical error that does not have strong impact on understanding, e.g. \textit{what some good book for read};
          \item 2: Can not understand what it means but it is still in the form of human language, e.g. \textit{what is the best movie of movie};
          \item 1: Non-sense composition of words and not in the form of human language, e.g. \textit{how world war iii world war}.
        \end{itemize}
      \item Coherent and Consistent
        \begin{itemize}
          \item 5: Accurate paraphrase with exact the same meaning of the source sentence;
          \item 4: Basically the same meaning of the source sentence but does not cover some minor content, e.g. \textit{what are some good places to visit in hong kong during summer $\rightarrow$ can you suggest some places to visit in hong kong};
          \item 3: Cover part of the content of source sentence and has serious information loss, e.g. \textit{what is the best love movie by wong ka wai $\rightarrow$ what is the best movie};
          \item 2: Topic relevant but fail to cover most of the content of source sentence, e.g. \textit{what is some tips to learn english $\rightarrow$ when do you start to learn english};
          \item 1: Topic irrelevant or even can not understand what it means.
        \end{itemize}
    \end{itemize}

There is token \textit{[UNK]} that stands for unknown token in paraphrase. Ones that contains \textit{[UNK]} should have both grammar and coherent score lower than 5. The grammar score should depend on other tokens in the paraphrase. The specific coherent score depends on the impact of \textit{[UNK]} on that certain paraphrase. Here are some paraphrase examples given original question \textit{how can robot have human intelligence ?}:
    \begin{itemize}
        \item \textit{paraphrase:} how can [UNK] be intelligent ? \\
            \textit{coherent score:} 1\\
             This token prevent us from understanding the
            question and give proper answer. It causes serious information loss here;
        \item \textit{paraphrase:} how can robot [UNK] intelligent ?\\
                \textit{coherent score:} 3\\
             There is information loss, but the unknown token does not influence our understanding so much;
        \item \textit{paraphrase:} how can robot be intelligent [UNK] ?\\
                \textit{coherent score:} 4\\
             \textit{[UNK]} basically does not influence understanding.
    \end{itemize}

NOTED:
    \begin{itemize}
      \item Please decouple \textit{grammar} and \textit{coherent} as possible as you can. For instance, given a sentence \textit{is it true that girls like shopping}, the paraphrase \textit{do girls like go go shopping} can get a \textit{coherent} score of 5 but a \textit{grammar} score of only 3. But for the one you even can not understand, e.g., \textit{how is the go shopping of girls}, you should give both of low \textit{grammar} score and low \textit{coherent} score, even it contains some topic-relevant words.
      \item Do a Google search when you see any strange entity name such that you can make more appropriate judgement.
    \end{itemize}

\end{document}